\documentclass{bmvc2k}

\usepackage{booktabs}
\usepackage{multirow}
\usepackage{makecell}
\usepackage{hyperref}

\title{Referring Expression Object Segmentation with Caption-Aware Consistency}

\addauthor{Yi-Wen Chen}{chenyiwena@gmail.com}{1}
\addauthor{Yi-Hsuan Tsai}{ytsai@nec-labs.com}{2}
\addauthor{Tiantian Wang}{tiantianwang.ice@gmail.com}{3}
\addauthor{Yen-Yu Lin}{yylin@citi.sinica.edu.tw}{1}
\addauthor{Ming-Hsuan Yang}{mhyang@ucmerced.edu}{34}

\addinstitution{
Academia Sinica
}
\addinstitution{
NEC Laboratories America
}
\addinstitution{
University of California, Merced
}
\addinstitution{
Google Cloud
}

\runninghead{Y.-W. Chen \MakeLowercase{et al}.}{Referring Expression Object Segmentation}

\def\ie{\emph{i.e}\bmvaOneDot}
\def\eg{\emph{e.g}\bmvaOneDot}

\def\etal{\emph{et al}\bmvaOneDot}

\begin{document}

\maketitle

\begin{abstract}
Referring expressions are natural language descriptions that identify a particular object within a scene and are widely used in our daily conversations. 
In this work, we focus on segmenting the object in an image specified by a referring expression.
To this end, we propose an end-to-end trainable comprehension network that consists of the language and visual encoders to extract feature representations from both domains.
We introduce the spatial-aware dynamic filters to transfer knowledge from text to image, and effectively capture the spatial information of the specified object.
To better communicate between the language and visual modules, we employ a caption generation network that takes features shared across both domains as input, and improves both representations via a consistency that enforces the generated sentence to be similar to the given referring expression.
%
We evaluate the proposed framework on two referring expression datasets and show that our method performs favorably against the state-of-the-art algorithms.

\end{abstract}

\section{Introduction}

Object segmentation aims to separate foreground objects from the background.
%
%
In this work, we focus on object segmentation from referring expressions, in which the segmentation is guided by a natural language description that identifies a particular object instance in a scene, \eg, {\em the man in a blue jacket} or {\em the laptop on the left}.
%
%
\begin{figure}[htp]
	\centering
	\includegraphics[width=0.65\linewidth]{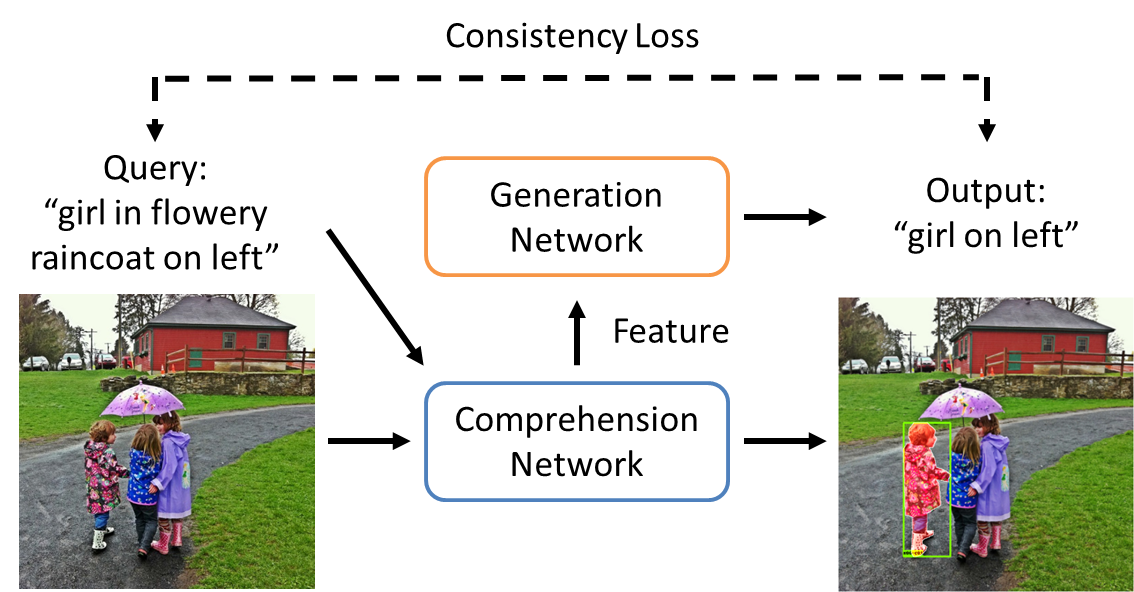}
	\caption{Illustration of the proposed algorithm. Given an image and a referring expression, we use a comprehension network to segment the specified object. With the features containing both language and visual information as input, the generation network produces a sentence identifying the target object. By enforcing a caption-aware consistency loss between the query and output sentence, we further improve the performance of comprehension.}
	\label{fig:overview}
	\vspace{-3mm}
\end{figure}
%
%
%
%
%

%
Transferring knowledge between language and visual domains is an important but challenging task.
%
Two relevant tasks are: 1) {\em referring expression comprehension} for localizing or segmenting an object according to a natural language description, and 2) {\em referring expression generation} for producing a sentence that identifies a particular object in an image.
%
Existing methods~\cite{Mao_CVPR_2016,Yu_ECCV_2016} address both tasks by constructing a generation model and inferring the region which maximizes the expression posterior in the comprehension task. 
However, such joint information is usually exploited to only enhance the generation performance.
%

%
%
%

In this paper, we focus on referring expression object segmentation. 
Unlike existing methods, our model jointly considers both tasks to benefit the comprehension task. 
%
Intuitively, when one signal, \eg, a sentence, is transferred from the language domain to the visual domain, and then transferred back to the language domain, the transferred-back signal is supposed to be similar to the original one.
By exploiting this property, we develop a network that jointly considers referring expression comprehension and generation, and enforces a caption-aware consistency between the visual and language domains.
%

%
%
%

%

To this end, we first design a comprehension network that contains the language and visual encoders to extract the feature representations of respective domains.
To connect these two domains, we further propose to use {\em spatial-aware dynamic filters} to bridge the language and visual encoders.
Meanwhile, these filters provide visual representations with the localization ability from the input referring expression.
Based on the proposed baseline model, we then employ a caption generation model that takes feature representations from the comprehension network as inputs.
%
The generated referring expression should be similar to the original sentence, and we leverage this property as an additional consistency cue to enhance the language and visual representations.
%
The main steps of the proposed model are illustrated in Figure~\ref{fig:overview}.

%
%
%
%

To evaluate the proposed method, we conduct extensive experiments on the RefCOCO \cite{Yu_ECCV_2016} and RefCOCOg~\cite{Mao_CVPR_2016,Nagaraja_ECCV_2016} datasets.
Experimental results show that our model performs favorably against the state-of-the-art methods.
In addition, we provide the ablation study to demonstrate the effectiveness of each component in the proposed framework, including the spatial-aware dynamic filters and caption-aware consistency.
%
%
%
%
The main contributions of this work are summarized as follows:
1) We integrate referring expression generation into referring expression comprehension so that the two complementary tasks can benefit each other via enforcing the caption-aware consistency.
2) We develop the spatial-aware dynamic filters that bridge the visual and language domains and facilitate the feature learning process.
3) We design an end-to-end trainable network for referring expression comprehension, achieving the state-of-the-art performance.
%

%
%
%
%




\section{Related Work}
{\flushleft {\bf Referring Expression Comprehension.}}
The task of referring expression comprehension aims to localize or segment an object given a natural language description.
Existing methods~\cite{Hu_CVPR_2016_2,Luo_CVPR_2017,Mao_CVPR_2016} mainly rely on recurrent caption generation models, and select the object with the maximum posterior probability of the expression among all object proposals.
By exploring the relationship between the object and its context~\cite{Nagaraja_ECCV_2016,Yu_ECCV_2016,Zhang_CVPR_2018}, the target object can be better localized.
%
%
%
%
Recent approaches adopt various learning strategies, such as embedding images and sentences into a common feature space~\cite{Wang_CVPR_2016,Rohrbach_ECCV_2016},
or learning attributes~\cite{Liu_ICCV_2017_2} to help differentiate objects of the same category.
%
%
In addition, Hu~\etal~\cite{Hu_CVPR_2017} analyze the inter-object relationships by parsing the sentence into subject, relationship and object parts.
%
%
To jointly consider the associated factors such as attributes and relationships between objects, Yu~\etal~\cite{Yu_CVPR_2018} propose a modular attention network to decompose the expression into subject appearances, locations, and relationships to other objects.
%

%
%
%
%
%
%
%
%

%
While the aforementioned methods mainly localize an object by a bounding box, algorithms that focus on segmentation~\cite{Hu_ECCV_2016,Li_CVPR_2018,Liu_ICCV_2017,Shi_ECCV_2018,Margffoy-Tuay_ECCV_2018} usually encode the referring expression through the LSTM network and use a fully convolutional network for foreground/background segmentation by using both the language and visual features.
Different from these approaches, our proposal-based model first localizes objects and performs segmentation via learning better feature representations through a referring generation network that considers the caption consistency.
%
%
We note that the approach in \cite{Rohrbach_ECCV_2016} also considers the consistency between the generated sentence and input sentence but does not target at segmenting objects.
Furthermore, this approach uses pre-defined and fixed region proposals, in which the visual representations are not updated through the proposals.
In contrast, our unified framework is end-to-end trainable while bridging features across visual and language domains.

{\flushleft {\bf Referring Expression Generation.}}
The generation task of referring expressions is a special case of image captioning. 
Rather than describing the whole image, the generated sentence uniquely identifies an object within the image.
A referring expression is considered good if one can localize the corresponding object by comprehending this referring expression.
Therefore, referring expression comprehension is often employed in the generation task~\cite{Mao_CVPR_2016,Liu_ICCV_2017_2,Luo_CVPR_2017} to improve the performance.
%

%
%
%
%

CNN-LSTM based models are widely used for image captioning~\cite{Karpathy_CVPR_2015,Rennie_CVPR_2017,Vinyals_CVPR_2015,Xu_ICML_2015}.
While a CNN model extracts visual features, an LSTM module produces captions.
To address referring expression generation, Mao~\etal~\cite{Mao_CVPR_2016} combine the extracted visual features with the location and size of the target object.
Furthermore, this method uses a CNN-LSTM model for the comprehension task and jointly trains the generation and comprehension modules.
%
%
%
%
%
Yu~\etal~\cite{Yu_CVPR_2017} further propose a joint speaker-listener-reinforcer model where a reward function is introduced to guide the expression sampling.
%
Their approach jointly trains the generation and comprehension networks, but does not specifically consider the caption consistency as our framework.
%
While the aforementioned methods mainly utilize the comprehension model to generate high-quality sentences, in this work, we focus on the comprehension task and demonstrate that the generation model also facilitates the comprehension performance by enforcing the proposed caption-aware consistency between the visual and language domains. 
%
%
%
%
%
%
%
%
%
\begin{figure}[t]
	\centering
	\includegraphics[width=0.9\linewidth]{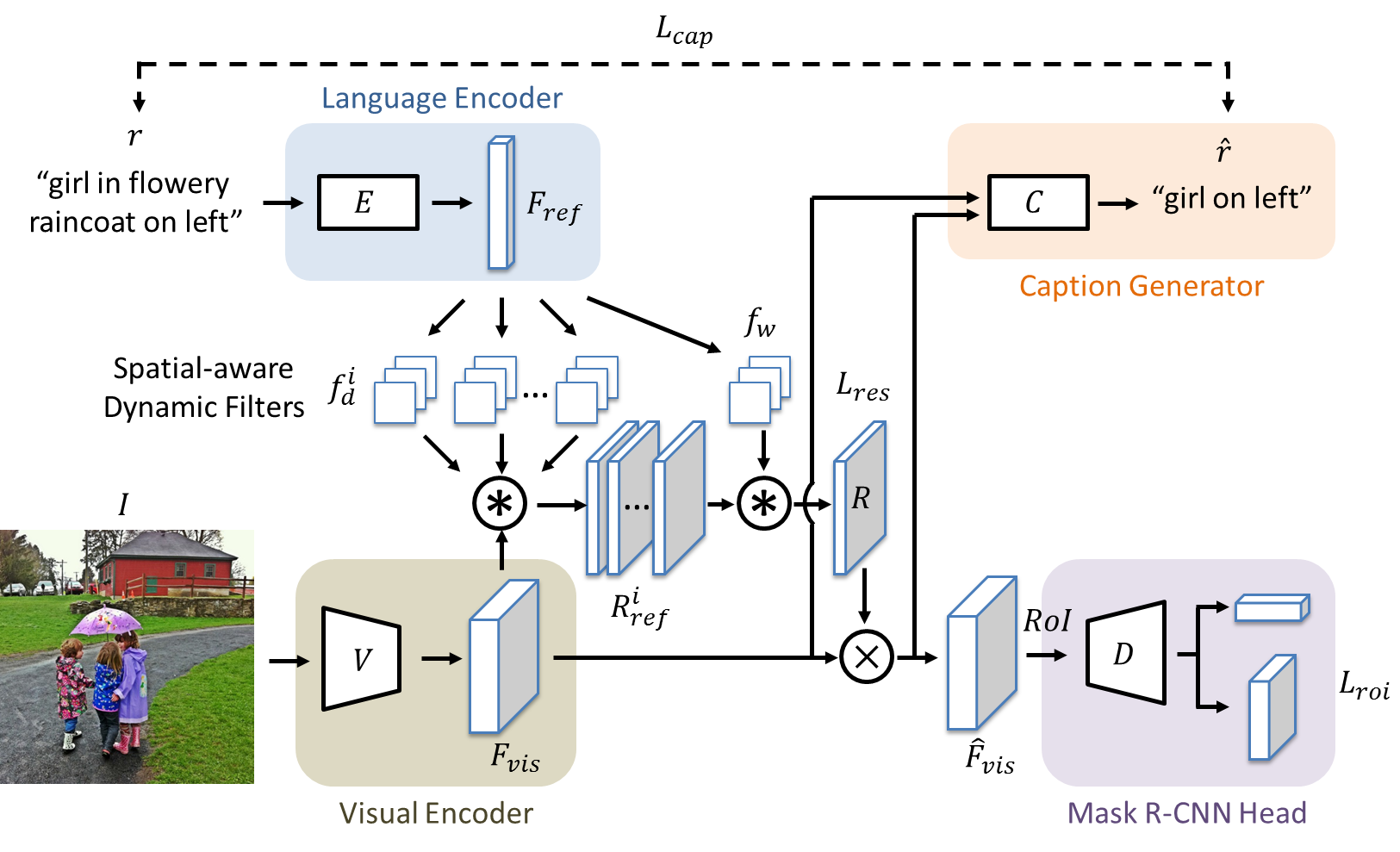}
	\caption{Architecture of the proposed framework. 
	The proposed network is composed of a language encoder $E$, a visual encoder $V$, a Mask R-CNN head $D$, and a caption generator $C$. 
	Features of the referring expression $r$ and image $I$ are extracted by $E$ and $V$ respectively. 
	Knowledge is transferred from language domain to visual domain via the spatial-aware dynamic filters $f^i_d$, by which a response map $R$ is generated to produce a location-aware feature $\hat{F}_{vis}$. 
	Based on this feature $\hat{F}_{vis}$ that carries information from both domains, we generate object bounding box and mask by the Mask R-CNN head $D$. The caption generator takes features $F_{vis}$ and $\hat{F}_{vis}$ as inputs, and produces a sentence identifying the target object. 
	We apply a caption-aware consistency loss between the input query $r$ and the generated sentence $\hat{r}$ to further improve the language and visual feature representations.}
	\label{fig:framework}
	\vspace{-3mm}
\end{figure}

\section{Proposed Framework}
In this work, we focus on referring expression object segmentation. 
The overview of the proposed framework is illustrated in Figure~\ref{fig:framework}.
Given an image $I$ and a natural language description $r$, we aim to segment the object in $I$ specified by $r$.
To this end, we propose an end-to-end trainable network that contains a language encoder $E$, a visual encoder $V$, a Mask R-CNN head $D$, and a caption generator $C$.
%
%
The encoders $E$ and $V$ extract language and visual features, respectively. 
Motivated by the dynamic filter network~\cite{Brabandere_NIPS_2016}, we enhance the ability of specific object localization via introducing the {\em spatial-aware dynamic filters} to transfer knowledge from text to image.
The yielded cross-modal information allows the Mask R-CNN head $D$ to produce more accurate segmentation results.
%
%
%
To further improve our model, we employ the caption generation network $C$ and a consistency loss $L_{cap}$ to jointly train the comprehension and generation networks.
%
%
%
We describe each component of the proposed network below.

%
%
%

\subsection{Segmentation from Referring Expression}
In this subsection, we introduce how the proposed network generates the object segment given the query referring expression. 
To this end, the language encoder $E$, visual encoder $V$, and spatial-aware dynamic filters are elaborated.
%
%
\label{sec:seg}
{\flushleft {\bf Language Encoder.}}
Similar to~\cite{Yu_CVPR_2018}, we use a bi-directional LSTM model to extract features of a referring expression. 
Given a referring expression $r = \{w_t\}_{t=1}^T$ of $T$ words with each word $w_t$ represented by a one-hot vector $e_t$, the bi-directional LSTM $S$ is applied to encode the whole sentence in both forward and backward directions:
%
\begin{align}
\overrightarrow{h}_t = \overrightarrow{S}(e_t, \overrightarrow{h}_{t-1}) \notag\\
\overleftarrow{h}_t = \overleftarrow{S}(e_t, \overleftarrow{h}_{t+1}) \notag\\
F_{ref} = [\overrightarrow{h}_T, \overleftarrow{h}_1],
\label{eq:lstm}
\end{align}
where $\overrightarrow{h}_t$ and $\overleftarrow{h}_t$ are the forward and backward hidden states at time step $t$, respectively. 
We concatenate the final hidden states in both directions to yield the feature representation $F_{ref}$ of the referring expression.
%
%

{\flushleft {\bf Visual Encoder.}}
%
Given an input image $I$, we aim at pixel-wise segmentation.
Different from the approaches based on the fully convolutional network (FCN) that does not generate instance-aware results, we adopt the proposal-based Mask R-CNN~\cite{He_ICCV_2017} framework to generate an object mask based on each detected object bounding box.
We use the ResNet-101~\cite{He_CVPR_2016} model as the backbone network and extract features over the entire image. 
The feature from the final convolutional layer of the fourth block, denoted by $F_{vis} = V(I)$, serves as the representation of image $I$.
%
%
%
%
%

{\flushleft {\bf Spatial-aware Dynamic Filters.}}
%
%
Motivated by the recent work~\cite{Li_CVPR_2017} on tracking with natural language, we utilize dynamic convolutional filters as a bridge to connect the language and visual domains.
%
%
%
%
Unlike conventional convolutional filters that apply the same weights to all input images, dynamic convolutional filters are generated depending on the input sentence.
Given the feature representation $F_{ref}$ of a sentence $r$, a single fully connected layer parameterized by the weights $W_d^1$ and the bias $b_d^1$ is adopted to generate a set of dynamic filters:
\begin{equation}
f_d^1 = \tanh(W_d^1 \cdot F_{ref} + b_d^1),
\label{eq:filter}
\end{equation}
where $\tanh$ is the hyperbolic tangent function, and $f_d^1$ is a set of $1 \times 1$ convolutional filters with the same number of channels as the visual representation $F_{vis}$. 
We then convolve the visual representation $F_{vis}$ with the generated dynamic filters $f_d^1$ to obtain a response map $R_{ref}^1$:
%
\begin{equation}
R_{ref}^1 = f_d^1 * F_{vis}.
\label{eq:map}
\end{equation}
With this formulation, knowledge is transferred from the language domain through learning the dynamic filters, with which the response map reflects the information inferred from the referring expression.
%
%

However, such filters consider the entire image and thus may only be able to catch the global structure but ignore spatially distributed objects.
As such, we propose to utilize spatial-aware dynamic convolutional filters that consider local regions of the image, including up, down, left, right, horizontal and vertical middle regions, and each region covers a half area of the entire image.
We thereby apply six additional fully connected layers to generate spatial-aware dynamic filters 
$\{f_d^i\}_{i=2}^7$
corresponding to each region $i$ via \eqref{eq:filter}. The six dynamic filters are then convolved with the visual feature $F_{vis}$, where the values outside the defined regions are set to 0. Then we obtain six spatial-aware response maps similar to \eqref{eq:map}, denoted by $\{R^i_{ref}\}_{i=2}^7$, in which each map focuses on its defined region.

To combine these spatial response maps and the one from \eqref{eq:map}, we adopt another set of dynamic filters $f_w$ with 7 channels, which are also generated from the sentence representation $F_{ref}$, to account for the importance of each region depending on the input sentence.
We convolve $f_w$ with the concatenation of the $7$ response maps $R_{con} = \text{concat}(R^i_{ref})$ and obtain a final response map $R$ with one channel, \ie,
\begin{equation}
R = \sigma(f_w * R_{con}),
\label{eq:attn}
\end{equation}
where $\sigma$ is the sigmoid function with output range $[0, 1]$.
Ideally, $R$ represents a map of the object specified by the input referring expression.
Thus, we apply a binary cross-entropy loss $L_{res}$ to supervise the response map $R$ with respect to the ground-truth object mask.
%
%
%
{\flushleft {\bf Baseline Objective.}}
%
%
Based on the response map in \eqref{eq:attn}, we take the element-wise multiplication of $R$ and $F_{vis}$ to be the caption-aware feature representation $\hat{F}_{vis}$, which carries the information from both the language and visual domains.
To obtain the final segmentation result, we then feed $\hat{F}_{vis}$ into the Mask R-CNN~\cite{He_ICCV_2017} RoI head $D$, which includes the bounding box and the binary segmentation branches. The overall objective can be written as:
\begin{equation}
L = L_{roi} + L_{res},
\label{eq:loss_seg}
\end{equation}
where $L_{roi}$ includes the classification loss, bounding box loss and mask loss, the same as 
those
defined in Mask R-CNN.
%
%
%
With this formulation, we construct an end-to-end trainable network that produces the referring expression object segmentation.
Unlike the state-of-the-art methods, such as MAttNet~\cite{Yu_CVPR_2018}, that require multiple training stages and pre-processing steps, our model can be efficiently learned, through the help of spatial-aware dynamic filters which provide the spatial information from the input sentence.
\subsection{A Joint Framework}
\label{sec:cap}
%
%
In light of the cycle consistency work~\cite{Zhu_ICCV_2017} that solves the domain transfer problem in cross-directions, we integrate both the referring expression comprehension and generation tasks into a joint framework, where their feature representations are shared and can be jointly optimized through back-propagation.
{\flushleft {\bf Referring Expression from Segmentation.}}
To generate a sentence describing a particular object within an image, we adopt the attention-based image captioning model~\cite{Xu_ICML_2015}.
%
%
%
To train the caption generation model $C$, we input the feature representation $F_{vis}$ extracted from Mask R-CNN and concatenate it with $\hat{F}_{vis}$ which contains the spatial information about the object.
%
%
As a result, during training the caption generation model, gradients can be back-propagated through $F_{vis}$ to update the Mask R-CNN feature extractor, as well as through $\hat{F}_{vis}$ to optimize dynamic filters and the language encoder.
%

{\flushleft {\bf Caption-aware Consistency.}}
Given the ground-truth sentence $r = \{w_t\}_{t=1}^T$, which is the input to the language encoder, the objective for caption generation is to minimize the cross-entropy loss $L_{cap}$:
\begin{equation}
L_{cap} = - \sum_{t=1}^T \log(p_{\theta_c}(w_t | w_1, ..., w_{t-1})),
\label{eq:loss_cap}
\end{equation}
where $p_{\theta_c}(\hat{w}_t | \hat{w}_1, ..., \hat{w}_{t-1})$ is the probability of predicting a particular word from the caption generation network parameterized by $\theta_c$.
Here, this loss function in our framework enforces that the predicted sentence $\hat{r}$ generated by the feature $\hat{F}_{vis}$, \ie, $\hat{r} = C(\hat{F}_{vis}, \cdot)$, should be consistent with the input query $r$ that generates the same feature, \ie, $\hat{F}_{vis} = \mathcal{F}(E(r))$, where $\mathcal{F}$ is a mixed operation involving the visual encoder $V$ and dynamic filters in the proposed method.
Hence, our caption-aware consistency actually enforces $r \approx C(\mathcal{F}(E(r)), \cdot)$.
{\flushleft {\bf Overall Objective.}}
%
\label{sec:full}
To exploit the caption-aware consistency, we jointly train the comprehension model, including the language encoder $E$, visual encoder $V$, Mask R-CNN head $D$ in Section~\ref{sec:seg} and the caption generation model $C$ in Section~\ref{sec:cap}.
The total loss function is extended from \eqref{eq:loss_seg} to:
%
\begin{equation}
L = L_{roi} + L_{res} + \alpha L_{cap},
\label{eq:loss_total}
\end{equation}
where $\alpha$ is the coefficient of the consistency loss.
We note that adding $L_{cap}$ enables the joint optimization between the language and the visual domains.
That is, the intermediate feature $\hat{F}_{vis}$ would be updated by the guidance from the first two loss functions in \eqref{eq:loss_total}, which are supervised by the comprehension task, and in the meanwhile $L_{cap}$ updates the feature based on the caption generation task.
%
%
%
\begin{table}[!t]
	\centering
	\footnotesize
	\def\arraystretch{0.9}
	\setlength\tabcolsep{5pt}
	\caption{Localization results of our method and the competing methods on two datasets. We summarize the major information used in each method, including context (C), attribute prediction (Attr), attention module (Attn), location (L), relationships between objects (R), and joint training with referring expression generation (J).}
	\vspace{2mm}
	\label{tab:det}
	\begin{tabular}{@{}@{}c c c c c c c c@{}}
		\toprule
		& \multicolumn{1}{c}{} & \multicolumn{3}{c}{RefCOCO} & \multicolumn{3}{c}{RefCOCOg} \\
		\cmidrule(lr){3-5} \cmidrule(lr){6-8}
		Model & Info. & val & testA & testB & val* & val & test \\ \midrule
		Nagaraja~\etal~\cite{Nagaraja_ECCV_2016} & C & 57.30 & 58.60 & 56.40  & -  & - & 49.50 \\
		Luo~\etal~\cite{Luo_CVPR_2017} & J & - & 67.94 & 55.18 & 49.07 & - & - \\
		Liu~\etal~\cite{Liu_ICCV_2017_2}& Attr, J & - & 72.08 & 57.29 & 52.35 & - & - \\
		Yu~\etal~\cite{Yu_CVPR_2017} & J & - & 73.78 & 63.83 & 59.84 & - & - \\
		MAttNet~\cite{Yu_CVPR_2018} & Attr, Attn, L, R & 76.65 & 81.14 & 69.99 & - & 66.58 & 67.27 \\
		VC~\cite{Zhang_CVPR_2018} & C & - & 73.33 & 67.44 & 62.30 & - & - \\
		\midrule
		baseline & - & 72.65 & 76.65 & 65.75 & 54.18 & 58.09 & 58.32 \\
		+ spatial coords~\cite{Hu_ECCV_2016} & L & 75.89 & 78.57 & 68.54 & 61.37 & 64.10 & 64.21 \\
		+ spatial-aware filters & L & 76.98 & 79.30 & 69.75 & 61.65 & 65.18 & 65.28 \\
		+ caption-aware consistency & J & 76.05 & 78.84 & 69.36 & 60.69 & 64.71 & 63.79 \\
		full model & L, J & \textbf{77.08} & \textbf{80.34} & \textbf{70.62} & \textbf{62.34} & \textbf{65.83} & \textbf{65.44} \\
		\bottomrule
	\end{tabular}
	\vspace{-3mm}
\end{table}
\subsection{Model Training and Implementation Details}
To train the joint network model, we adopt a sequential training strategy to optimize the objective in \eqref{eq:loss_total}.
First, we only update the comprehension network by optimizing \eqref{eq:loss_seg}.
Then, we pre-train the caption generation network by optimizing \eqref{eq:loss_cap} as a warm-up.
Finally, we update the entire framework with the objective in \eqref{eq:loss_total}.
%
%
With the trained model, we choose the detected object with the largest score during testing.
%
%

We implement our model with PyTorch using the SGD optimizer.
%
For the language encoder, the dimension of the LSTM hidden states is set to $512$. 
By concatenating the forward and backward hidden states, the feature $F_{ref}$ is a $1024$-dimensional vector.
In the visual encoder, the visual feature $F_{vis}$ is of dimension $1024$. 
Thus, we also generate the dynamic filters of dimension $1024$.
For the caption generation model, the input spatial features are resized to $14 \times 14$ and have the same number of channels as that of the concatenation of $F_{vis}$ and $\hat{F}_{vis}$.
When training the full model, the loss weight $\alpha$ in \eqref{eq:loss_total} is set to $0.1$ for all experiments. 
The codes and models are available at: \url{https://github.com/wenz116/lang2seg}.
\section{Experimental Results}
We evaluate the proposed framework on two referring expression datasets: RefCOCO~\cite{Yu_ECCV_2016} and RefCOCOg (with two splits\footnote{The first split~\cite{Mao_CVPR_2016} randomly partitions objects into training and validation sets. We denote the validation set as ``val*'' in this paper. The second split~\cite{Nagaraja_ECCV_2016} randomly partitions images into training, validation and testing sets, where we denote the validation and testing ones as ``val'' and ``test'', respectively.})~\cite{Mao_CVPR_2016,Nagaraja_ECCV_2016}.
The two datasets are collected from the Microsoft COCO images~\cite{MSCOCO}, with different properties of expressions.
%
%
We show both detection and segmentation results with comparisons against the state-of-the-art algorithms.
In addition, we present an ablation study to demonstrate the importance of each component in the proposed framework.
%
More results are provided in the supplementary material.

For evaluating the detection performance, the predicted bounding box is considered correct if the intersection-over-union (IoU) of the prediction and the ground truth is above 0.5. As for the segmentation quality, we use Intersection-over-Union (IoU) as metric.
%
%
\subsection{Localization Results}
In Table~\ref{tab:det}, we show comparisons with existing state-of-the-art algorithms~\cite{Liu_ICCV_2017_2,Luo_CVPR_2017,Nagaraja_ECCV_2016,Yu_CVPR_2018,Yu_CVPR_2017,Zhang_CVPR_2018}.
Since each method adopts diverse information to help the comprehension task, we further summarize the major cues that each approach relies on, such as context information~\cite{Nagaraja_ECCV_2016,Zhang_CVPR_2018}, attribute prediction~\cite{Liu_ICCV_2017_2,Yu_CVPR_2018}, and joint training with referring expression generation~\cite{Liu_ICCV_2017_2,Luo_CVPR_2017,Yu_CVPR_2017}.

Table~\ref{tab:det} shows that the proposed method performs favorably against most methods by significant margins, and competitively with MAttNet~\cite{Yu_CVPR_2018}.
We note that, the MAttNet method utilizes various cues, including attention module, attribute prediction, location information, and relations between objects to achieve good performance, while our model only focuses on the location cue and joint training with referring expression generation.
It is also worth mentioning that our model is a unified framework that is end-to-end trainable, while MAttNet requires multiple separate training stages to obtain the final model.
%
The runtime speed of our method is 0.17 seconds per image, which is much faster than MAttNet with 0.67 seconds per image on 
an Intel Xeon 2.5 GHz machine and an NVIDIA GTX 1080 Ti GPU with 11 GB memory.
%
%
%
%
\begin{table*}[!t]
    \footnotesize
	\centering
	\def\arraystretch{0.9}
	\setlength\tabcolsep{5pt}
	\caption{Segmentation results of our method and the competing methods on two datasets.}
	\vspace{2mm}
	\label{tab:seg}
	\begin{tabular}{@{}@{}c c c c c c c c@{}}
		\toprule
		& \multicolumn{1}{c}{} & \multicolumn{3}{c}{RefCOCO} & \multicolumn{3}{c}{RefCOCOg} \\
		\cmidrule(lr){3-5} \cmidrule(lr){6-8}
		Model & Backbone Net & val & testA & testB & val* & val & test \\ \midrule
		D+RMI+DCRF~\cite{Liu_ICCV_2017} & Deeplab101 & 45.18 & 45.69 & 45.57 & - & - & - \\
		RRN+LSTM+DCRF~\cite{Li_CVPR_2018} & Deeplab101 & 55.33 & 57.26 & 53.95 & 36.45 & - & - \\
		MAttNet~\cite{Yu_CVPR_2018} & Res101 & 56.51 & 62.37 & 51.70 & - & 47.64 & 48.61 \\
		KWAN~\cite{Shi_ECCV_2018} & Deeplab101 & - & - & - & 36.92 & - & - \\
		DMN~\cite{Margffoy-Tuay_ECCV_2018} & DPN92 & 49.78 & 54.83 & 45.13 & 36.76 & - & - \\
		\midrule
		Ours & Res101 & 58.90 & 61.77 & 53.81 & 44.32 & 46.37 & 46.95 \\
		\bottomrule
	\end{tabular}
	\vspace{-3mm}
\end{table*}
\begin{figure}[!t]
    \footnotesize
	\centering
	\begin{tabular}
		{ @{\hspace{0mm}}c@{\hspace{1mm}} @{\hspace{0mm}}c@{\hspace{1mm}} @{\hspace{0mm}}c@{\hspace{1mm}} @{\hspace{0mm}}c@{\hspace{1mm}} @{\hspace{0mm}}c@{\hspace{1mm}} @{\hspace{0mm}}c@{\hspace{0mm}} }
		
		\makecell{``person holding tray''}  & \makecell{``girl in red coat''} & \makecell{``darkest colored horse''} & \makecell{``lighter brown horse\\ with head down''} \\

		\includegraphics[width=0.22\linewidth]{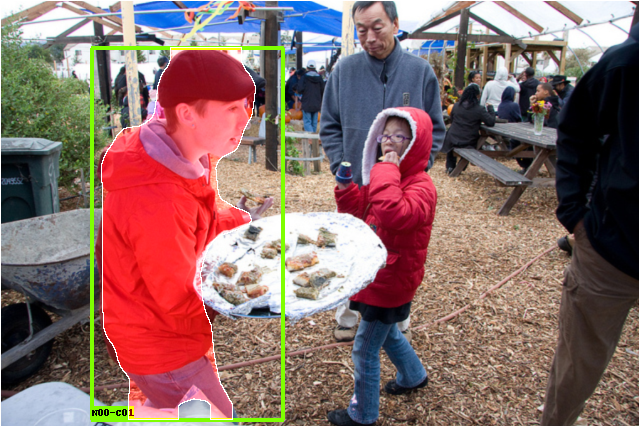} &
		\includegraphics[width=0.22\linewidth]{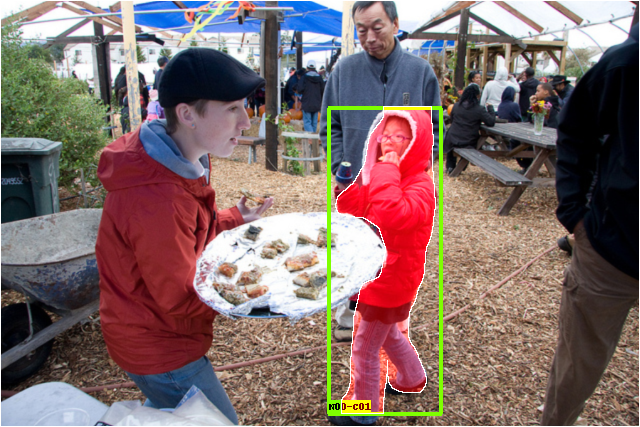} &

		\includegraphics[width=0.22\linewidth]{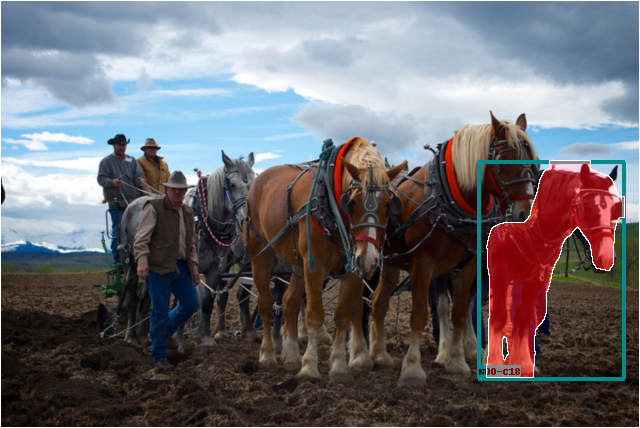} &
		\includegraphics[width=0.22\linewidth]{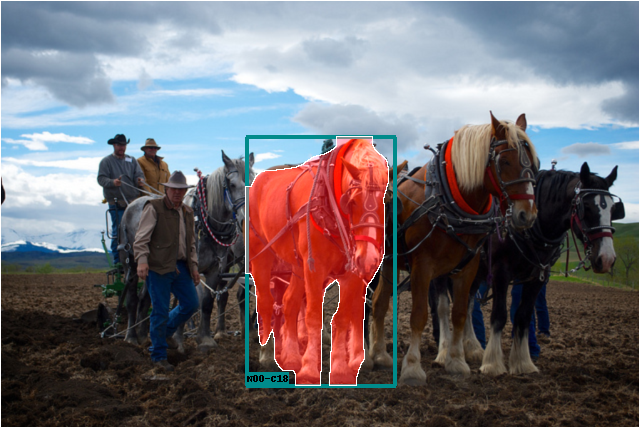} \\
		
		\makecell{``a small giraffe''}  & \makecell{``giraffe to the far left''} & \makecell{``the slice of cake\\ on the left''} & \makecell{``chocolate dessert cake\\ on a plate''} \\

		\includegraphics[width=0.22\linewidth]{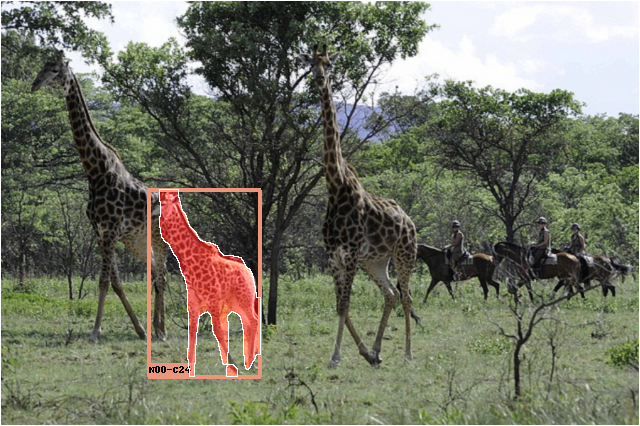} &
		\includegraphics[width=0.22\linewidth]{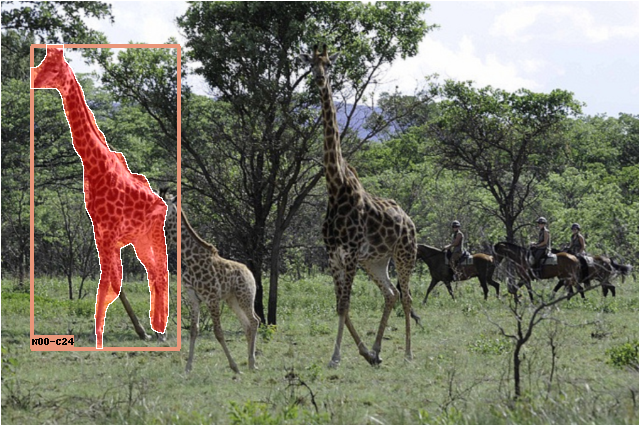} &

		\includegraphics[width=0.22\linewidth]{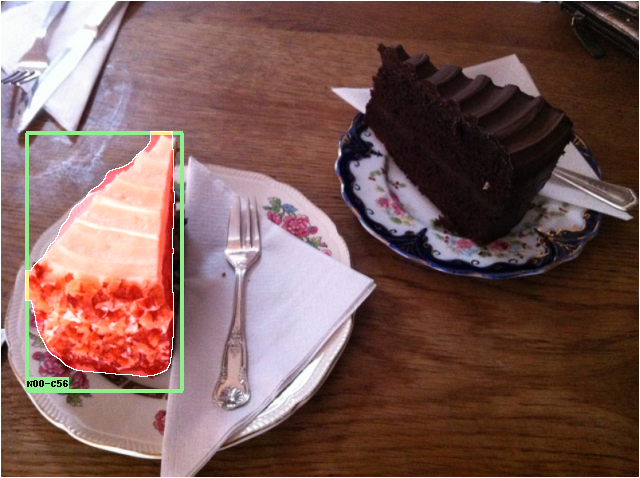} &
		\includegraphics[width=0.22\linewidth]{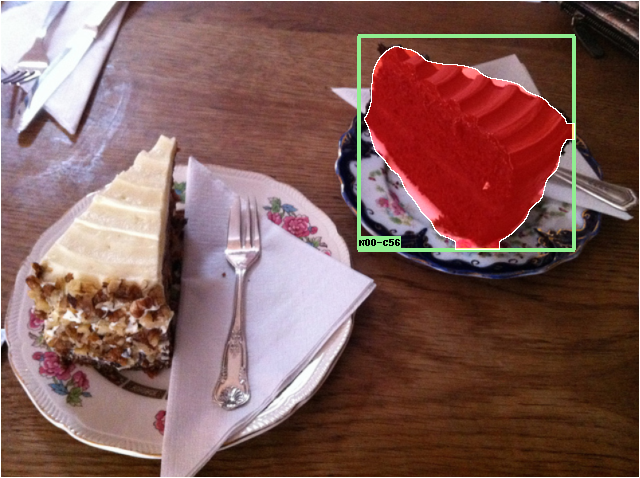} \\
	\end{tabular}
	\caption{Sample results of objects referred by various query expressions.}
	\label{fig:dif_obj}
	\vspace{-1mm}
\end{figure}
\begin{figure}[t]
    \footnotesize
	\centering
	\begin{tabular}
		{ @{\hspace{0mm}}c@{\hspace{1mm}} @{\hspace{0mm}}c@{\hspace{1mm}} @{\hspace{0mm}}c@{\hspace{1mm}} @{\hspace{0mm}}c@{\hspace{1mm}} @{\hspace{0mm}}c@{\hspace{1mm}} @{\hspace{0mm}}}

		``elephant in back'' \\
		\includegraphics[width=0.18\linewidth]{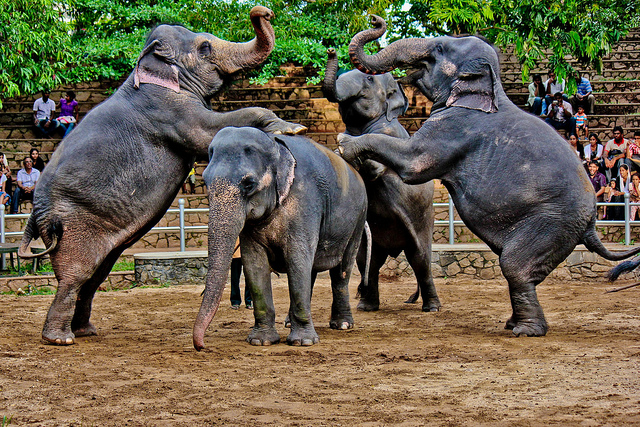} &	
		\includegraphics[width=0.18\linewidth]{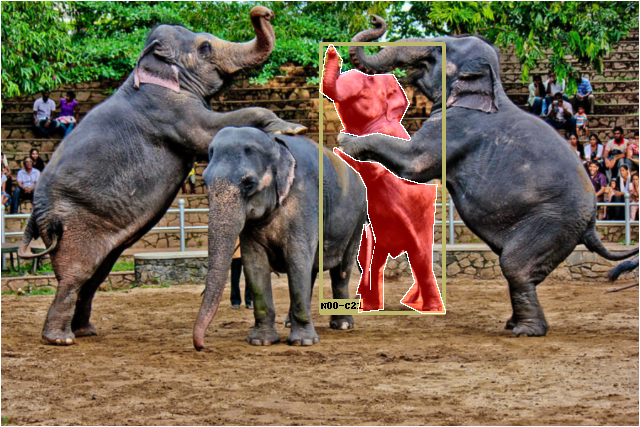} &
		
		\includegraphics[width=0.18\linewidth]{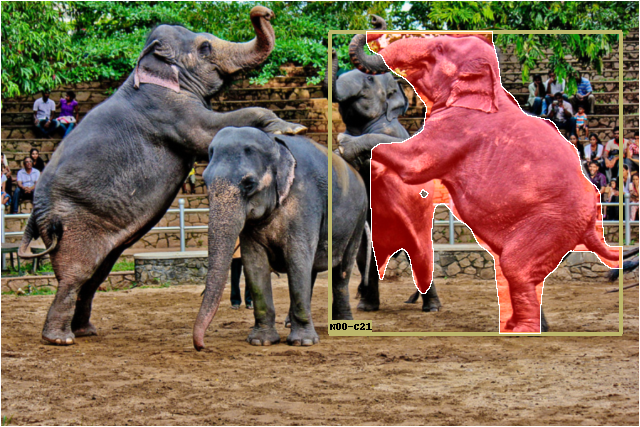} &
		
		\includegraphics[width=0.18\linewidth]{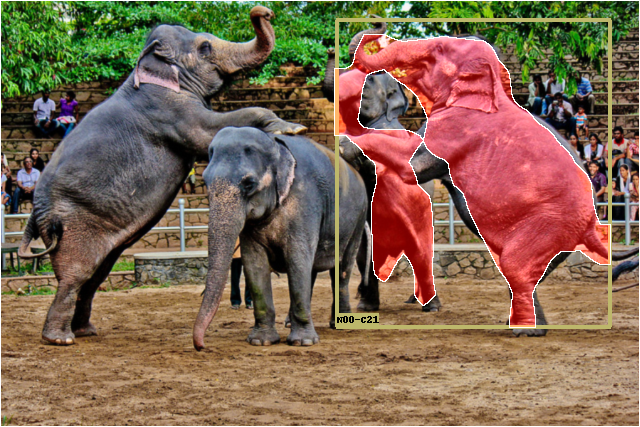} &
		\includegraphics[width=0.18\linewidth]{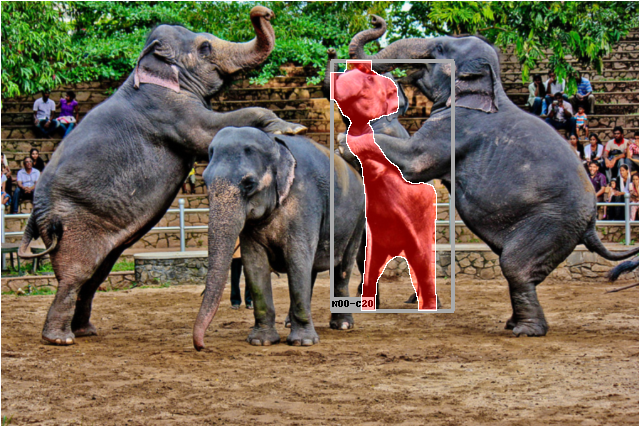} \\
		Input Image & Ground Truth & Baseline & Spatial-aware Filters & Full Model \\
		
	\end{tabular}
	\caption{Sample results from different variants of the proposed model on RefCOCO.}
	\label{fig:ablation}
    \vspace{-1mm}
\end{figure}
%
%
\subsection{Segmentation Results}
We present the experimental results with comparisons to the state-of-the-art algorithms including D+RMI+DCRF~\cite{Liu_ICCV_2017}, RRN+LSTM+DCRF~\cite{Li_CVPR_2018}, MAttNet~\cite{Yu_CVPR_2018}, KWAN~\cite{Shi_ECCV_2018} and DMN~\cite{Margffoy-Tuay_ECCV_2018} on the two datasets in Table~\ref{tab:seg}.
Overall, our method consistently and significantly outperforms other segmentation-based approaches that use a similar backbone network (\ie, Deeplab~\cite{CP2016Deeplab} with ResNet-101) as ours.
%
Different from the DMN~\cite{Margffoy-Tuay_ECCV_2018} scheme that utilizes dynamic filters in a sequential manner for capturing the information of each word in a sentence, our model generates the dynamic filters in a spatial-aware manner, where each set of filters produces a response map to certain region of the image.
We note that the proposed method achieves better performance.
%
Similar to the localization results, MAttNet~\cite{Yu_CVPR_2018} that fuses multiple cues performs competitively with our model.
%
%
%
We present qualitative examples of referring expression object segmentation in Figure~\ref{fig:dif_obj}. 
The proposed model can segment different objects according to various query expressions, such as the location, color, or action information, and further demonstrates the effectiveness of the proposed caption-aware consistency framework.
%
%
\subsection{Ablation Study}
%
%
We present the results of an ablation study in Table~\ref{tab:det}.
We first show that using the proposed spatial-aware dynamic filters improves the baseline with only a single dynamic filter or the spatial-aware mechanism~\cite{Hu_ECCV_2016} that concatenates spatial coordinates and feature maps.
%
%
Second, the referring expression generation network with caption-aware consistency performs favorably against the baseline model.
In the full model with both spatial-aware filters and caption-aware consistency, higher performance gains are achieved over other baselines.
%
%

We present sample segmentation results predicted by different variants of our model in Figure~\ref{fig:ablation}. 
Compared with the baseline and the model with spatial-aware filters, the proposed full model can localize objects accurately while the baseline model predicts the wrong object. 
In addition to improving the localizing ability, our full model enhances feature representations around the object. 
For instance, the elephant in back is well segmented by our model even if it is surrounded by  complex background and similar instances.
\section{Concluding Remarks}
In this paper, we propose an end-to-end trainable framework for referring expression segmentation.
We design a comprehension model that consists of language and visual encoders to extract feature representations in the respective domains. By introducing the spatial-aware dynamic filters, knowledge can be transferred from language domain to visual domain, while capturing the useful location cue.
In addition to the proposed baseline model, we employ a caption generation network to connect referring expression comprehension and generation.
Considering the consistency that the generated sentence is supposed to be similar to the given referring expression, we enforce a caption-aware consistency loss and further enhance the language and visual representations.
Extensive experiments and an ablation study on two referring expression datasets show that the proposed algorithm achieves favorable performance against the state-of-the-art methods.

{\flushleft {\bf Acknowledgments.}} This work was supported in part by Ministry of Science and Technology (MOST) under grants 107-2628-E-001-005-MY3 and 108-2634-F-007-009.

\bibliography{mybib}
\end{document}